\documentclass[letterpaper, 10pt, conference]{template/ieeeconf}
\def\TEMPOClassLoaded{1}
\AddToHook{begindocument/before}{%
  \definecolor{red}{rgb}{0,0,0}%
  \definecolor{qyg}{rgb}{0,0,0}%
  \definecolor{revisionblue}{rgb}{0,0,0}%
  \renewcommand{\qygdel}[1]{}%
  \renewcommand{\qygdelred}[1]{}%
  \hypersetup{hidelinks}%
}

\ifdefined\TEMPOClassLoaded
\else
\documentclass[letterpaper, 10pt, conference]{template/ieeeconf}  
\fi
\usepackage[T1]{fontenc}
\usepackage{graphicx}
\graphicspath{{template/figures/}{figures/}}
\newcommand{\realworldresultscaption}{%
\textbf{Real-world results} on (a) block sliding, (b) car retrieval, and (c) car placement.
Top: initial task states (Initial) and successful final states reached by TEMPO (Goal).
Bottom: online RL rewards for TEMPO and FLOWER-RL (mean $\pm$ SD, three seeds).%
}

\usepackage{wrapfig}
\usepackage{booktabs}
\usepackage{colortbl}
\usepackage{tabularx}
\usepackage{amsmath}
\usepackage{amsmath} 
\usepackage{amssymb}  
\usepackage[]{changes}
\presetkeys%
    {todonotes}%
    {inline,backgroundcolor=yellow}{}
\usepackage{chngcntr}
\usepackage{siunitx}
\usepackage{tikz}
\usepackage[outline]{contour}
\usepackage{adjustbox}
\usepackage{url}
\usepackage{subfigure}
\definecolor{resultgreen}{RGB}{232,246,232}
\definecolor{ourscurve}{RGB}{0,87,217}
\definecolor{flowerrlcurve}{RGB}{240,90,0}
\usepackage{ulem}
\usepackage{hyperref}

\definecolor{qyg}{RGB}{255, 0, 0}
\definecolor{revisionblue}{RGB}{0, 0, 255}

\newcommand{\qyg}[1]{\textcolor{qyg}{#1}}
\newcommand{\qygdel}[1]{\textcolor{revisionblue}{\sout{#1}}}
\newcommand{\qygdelred}[1]{{\color{qyg}\bgroup\markoverwith{\textcolor{revisionblue}{\rule[.5ex]{2pt}{.4pt}}}\ULon{#1}}}
\newcommand{\methodname}{TEMPO}

\makeatletter
\long\def\@makecaption#1#2{%
\ifx\@captype\@IEEEtablestring%
\begin{center}{\footnotesize #1: {\scshape #2}}\end{center}%
\@IEEEtablecaptionsepspace%
\else
\@IEEEfigurecaptionsepspace%
\setbox\@tempboxa\hbox{\footnotesize #1.~~ #2}%
\ifdim \wd\@tempboxa >\hsize%
\setbox\@tempboxa\hbox{\footnotesize #1.~~ }%
\parbox[t]{\hsize}{\footnotesize \noindent\unhbox\@tempboxa#2}%
\else%
\ifcenterfigcaptions \hbox to\hsize{\footnotesize\hfil\box\@tempboxa\hfil}%
\else \hbox to\hsize{\footnotesize\box\@tempboxa\hfil}%
\fi\fi\fi}
\makeatother

\IEEEoverridecommandlockouts                              

\title{\LARGE \bf
TEMPO: Semantic-Action Decoupled RL Post-Training for Vision-Language-Action Models
}
\author{Ziheng Liu$^1$, Quantao Yang$^{2, 3}$%
\thanks{$^1$School of Computer Science and Technology, Zhejiang Gongshang University, Hangzhou
310018, China.}%
\thanks{$^2$Department of Robotics, Perception, and Learning, KTH Royal Institute of Technology, Sweden. Contact: \texttt{quantao@kth.se}}%
\thanks{$^3$Qualia Robotics, Denmark.}%
}
\providecommand{\DIFaddtex}[1]{{\color{red}#1}} 
\providecommand{\DIFdeltex}[1]{} 
\providecommand{\DIFaddbegin}{} 
\providecommand{\DIFaddend}{} 
\providecommand{\DIFaddFL}[1]{\DIFadd{#1}} 
\providecommand{\DIFdelFL}[1]{} 
\providecommand{\DIFaddbeginFL}{} 
\providecommand{\DIFaddendFL}{} 
\providecommand{\DIFadd}[1]{\texorpdfstring{\DIFaddtex{#1}}{#1}} 
\makeatletter 
\let\sout@orig\sout 
\renewcommand{\sout}[1]{\ifmmode\text{\sout@orig{\ensuremath{#1}}}\else\sout@orig{#1}\fi} 
\makeatother 
\RequirePackage{listings} 
\lstdefinelanguage{DIFcode}{ 
  moredelim=[il][\color{white}\tiny]{\%DIF\ <\ }, 
  moredelim=[il][\sffamily\bfseries]{\%DIF\ >\ } 
} 
\lstdefinestyle{DIFverbatimstyle}{ 
	language=DIFcode, 
	basicstyle=\ttfamily, 
	columns=fullflexible, 
	keepspaces=true 
} 
\lstnewenvironment{DIFverbatim}[1][]{\lstset{style=DIFverbatimstyle,#1}}{} 
\lstnewenvironment{DIFverbatim*}[1][]{\lstset{style=DIFverbatimstyle,showspaces=true,#1}}{} 
\begin{document}

\maketitle
\thispagestyle{empty}
\pagestyle{empty}

\begin{abstract}
Vision-language-action (VLA) models are commonly adapted to downstream manipulation tasks via supervised fine-tuning (SFT) or online reinforcement learning (RL) post-training. SFT is prone to distribution mismatch, and existing RL approaches typically  \DIFaddbegin \DIFadd{optimize all trainable components within a single RL loop at a uniform update rate, although these components play }\DIFaddend distinct functional roles. We propose \methodname{}, a semantic-action decoupled, two-timescale RL post-training framework for VLA models. \methodname{} freezes the pretrained vision-language backbone  \DIFaddbegin \DIFadd{and adapts }\DIFaddend two components with dedicated RL  loops:
the semantic projection layer\DIFaddbegin \DIFadd{, which produces the latent action conditioning the action expert, }\DIFaddend and the low-level action expert  \DIFaddbegin \DIFadd{itself. The }\DIFaddend semantic projection layer  \DIFaddbegin \DIFadd{is updated infrequently }\DIFaddend and the action expert frequently,  \DIFaddbegin \DIFadd{which aims to limit latent-action drift while }\DIFaddend allowing the action expert to learn  from online feedback. \DIFaddbegin \DIFadd{Our analyses show that, compared with equal update frequencies, less frequent semantic updates reduce latent-action drift and improve policy performance.
}\DIFaddend Experiments on the CALVIN benchmark and real-world manipulation tasks demonstrate that \methodname{}  outperforms both pretrained state-of-the-art VLA models and the RL post-training baseline, while reaching and maintaining higher evaluation rewards on  \DIFaddbegin \DIFadd{three }\DIFaddend real-world tasks.
Project Page: \url{https://anonymous.4open.science/w/tempo-page/}.

\DIFaddbegin 

\DIFaddend \end{abstract}

\section{\uppercase{Introduction}}
\label{sec:introduction}

\DIFaddbegin 

\DIFadd{Pretrained Vision-Language-Action }\DIFaddend (VLA) models \DIFaddbegin \DIFadd{~\mbox{
\cite{black2025pi0, reuss2025flower} }\hskip0pt
}\DIFaddend are commonly adapted to downstream robotic tasks through supervised fine-tuning (SFT) on demonstration trajectories, using either full-parameter fine-tuning or parameter-efficient methods such as LoRA~\cite{kim2025openvla, kim2025openvlaoft}. Although effective, SFT is inherently limited by offline data: high-quality robot trajectories are expensive to collect and cannot exhaustively cover the states encountered during deployment. More fundamentally, SFT optimizes action prediction under the demonstration distribution rather than closed-loop success under the policy-induced state distribution. Small execution errors can therefore compound over long-horizon tasks and eventually lead to failure~\cite{ross2011dagger}. These limitations motivate online post-training paradigms that enable policies to adapt through task-specific feedback acquired from their own interactions with the environment.

\begin{figure}[t]
\centering
\makebox[\columnwidth][c]{\includegraphics[width=\columnwidth]{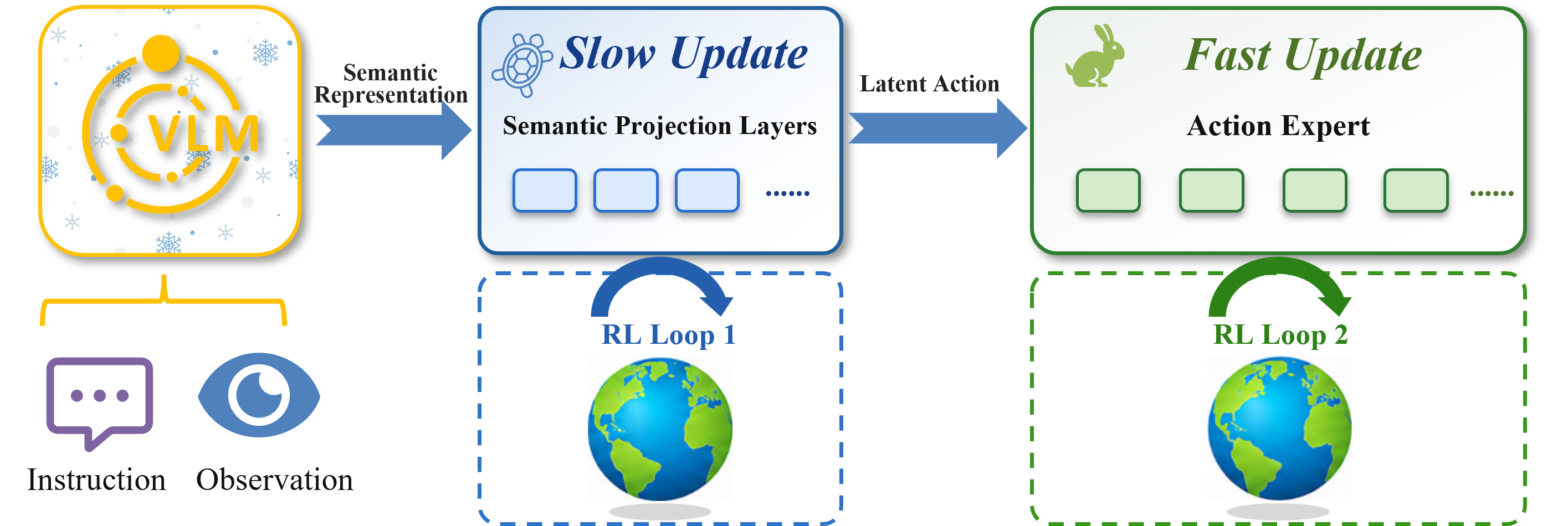}}
\caption{\textbf{Method overview.} \methodname{} keeps the VLM frozen and decouples the reinforcement
learning update frequencies of the semantic projection layer and action
expert, updating the former less frequently and the latter more frequently.}
\DIFaddbeginFL \vspace{-0.5cm}
\DIFaddendFL \label{fig:overview}
\end{figure}

Online reinforcement learning (RL)~\cite{lu2025vlarl, li2026simplevlarl, xu2026rl} directly optimizes task rewards through environment interaction, enabling VLA policies to learn from the outcomes of their own executions. However, existing methods typically optimize trainable components of a VLA 
\qyg{within a single RL loop and at a uniform update rate, although these components differ in their functional roles and adaptation requirements}. In modular VLA architectures, \qyg{a vision-language component produces task-relevant representations, which condition a separate action module that generates executable actions.}
\qyg{Prior work has separated semantic and action computation in other ways, for example by training an auxiliary RL policy on top of a frozen VLA~\cite{xu2026rl}, or alternating RL and supervised training stages~\cite{guo2025irevla}. However, to our knowledge, no prior work adapts the internal modules of a VLA through separate RL loops, and how often each module should be updated during online RL remains unexplored.}

On the semantic side, directly updating a large vision-language backbone not only increases computational cost but may also disturb its pretrained representations~\cite{driess2025knowledge,guo2025irevla}. We therefore preserve the internet-scale semantic knowledge encoded by the pretrained vision-language backbone and fine-tune the downstream semantic projection layer in a dedicated RL loop, as it still retains task-relevant information and serves as the semantic-side optimization target. On the action side, the action expert directly maps latent actions to executable action chunks, and we adapt it in a separate RL loop  \DIFaddbegin \DIFadd{using online }\DIFaddend control- and execution-related feedback .

After selecting the semantic projection layer and action expert as adaptation targets, a remaining question is how to coordinate their update frequencies. Frequent updates to the semantic projection layer may induce  \DIFaddbegin \DIFadd{latent-action }\DIFaddend drift, making the optimization target of the action expert continually shift and hindering stable learning. In contrast, the action expert must incorporate interaction feedback sufficiently quickly to improve control. A uniform update schedule may therefore be unable to balance these distinct adaptation requirements.

We therefore propose \methodname{}, a \underline{t}wo-timescale s\underline{em}antic-action decou\underline{p}led RL p\underline{o}st-training framework for VLA models, as illustrated in Fig.~\ref{fig:overview}.
\methodname{}
optimizes the semantic projection layer and the action expert
through two separate module-level TD3~\cite{fujimoto2018addressing} loops, while decoupling
their update frequencies: the semantic projection layer is updated less frequently, whereas the action expert is updated more frequently.
Our main contributions are summarized as follows:
\begin{itemize}
    \item We propose \methodname{}, a  \DIFaddbegin \DIFadd{semantic-action }\DIFaddend decoupled, two-timescale RL post-training framework that optimizes different VLA components through dedicated RL loops. Specifically, \methodname{} adapts the semantic projection layer while keeping the pretrained vision-language backbone frozen to preserve its large-scale semantic knowledge, and optimizes the action expert to learn control- and execution-related behaviors from online interaction feedback.
    \item We introduce a module-specific update-frequency strategy to coordinate the adaptation dynamics of semantic and action modules. \methodname{} updates the semantic projection layer less frequently to mitigate latent-action drift, while updating the action expert more frequently to efficiently incorporate control feedback.
    \item We evaluate \methodname{} on CALVIN and  \DIFaddbegin \DIFadd{three }\DIFaddend real-world manipulation tasks,  \DIFaddbegin \DIFadd{showing }\DIFaddend improved long-horizon performance over both state-of-the-art VLA models and the RL post-training baseline in simulation, as well as higher rewards in physical experiments.
\end{itemize}

\section{\uppercase{Related Work}}
\label{sec:related_work}

\subsection{\qyg{VLA Models and Supervised Fine-Tuning}}
\qyg{Scaling robot datasets~\cite{walke2023bridgedata} and generalist architectures~\cite{black2025pi0, li2023roboflamingo} have provided a strong foundation for adapting VLA policies to specific manipulation tasks.}
Building on these advances, supervised fine-tuning~\cite{chen2025visa} has become a common approach for adapting pretrained VLA policies to specific robots and tasks.
Such methods typically update model parameters using task-specific demonstrations by either full-parameter fine-tuning~\cite{kim2025openvla, fan2025diffusion} or LoRA-based adaptation~\cite{kim2025openvlaoft}.
These studies  \DIFaddbegin \DIFadd{show }\DIFaddend that supervised fine-tuning can effectively improve VLA policy performance on specific tasks  \DIFaddbegin \DIFadd{but is }\DIFaddend primarily driven by offline demonstration data.

Beyond general supervised fine-tuning strategies, recent studies have also begun to examine the interactions among functionally distinct VLA modules during adaptation. Knowledge insulation~\cite{driess2025knowledge} investigates gradient interference between the vision-language backbone and the action expert, isolating action-side gradients to reduce their interference with pretrained vision-language knowledge. However, it  \DIFaddbegin \DIFadd{relies on pre-collected training data without using feedback from the policy's own interactions for online improvement}\DIFaddend .

\subsection{Reinforcement Learning for VLA Post-Training}

To further improve pretrained VLA policies using online interaction feedback, existing RL post-training methods mainly focus on reward signals and optimization mechanisms.
For reward signals, VLA-RL~\cite{lu2025vlarl} uses a learned process reward, RIPT-VLA~\cite{tan2025riptvla} adopts sparse binary success feedback, and VLA-RFT~\cite{li2025vlarft} employs simulator-verified rewards. For optimization mechanisms,  different methods expose distinct but predetermined parameter sets to RL optimization. VLA-RL~\cite{lu2025vlarl} uses PPO to update the LoRA parameters of an autoregressive OpenVLA policy, whereas SimpleVLA-RL~\cite{li2026simplevlarl} applies GRPO to optimize all parameters of an action-token policy. iRe-VLA~\cite{guo2025irevla} keeps the vision-language backbone unchanged and updates only the action head during the RL stage, which is then alternated with supervised optimization over a broader set of model parameters.  \DIFaddbegin \DIFadd{RL Token~\mbox{
\cite{xu2026rl} }\hskip0pt
compresses VLA vision-language backbone features into state inputs for an auxiliary policy that refines reference action chunks through online RL, keeping the VLA frozen.
}

\DIFadd{Beyond the choice of trainable parameters, coordinating learning across modules has been studied in modular and hierarchical RL. Policy sketches~\mbox{
\cite{andreas2017sketches} }\hskip0pt
compose reusable subpolicies, whereas HIRO~\mbox{
\cite{nachum2018hiro} }\hskip0pt
coordinate high-level goals and lower-level control.
} \DIFadd{The coordination of components with distinct functional roles, including the timescales involved, remains less explored}\DIFaddend .
\begin{figure*}[t]
\vspace{0.2cm}
\centering
\includegraphics[width=0.98\textwidth]{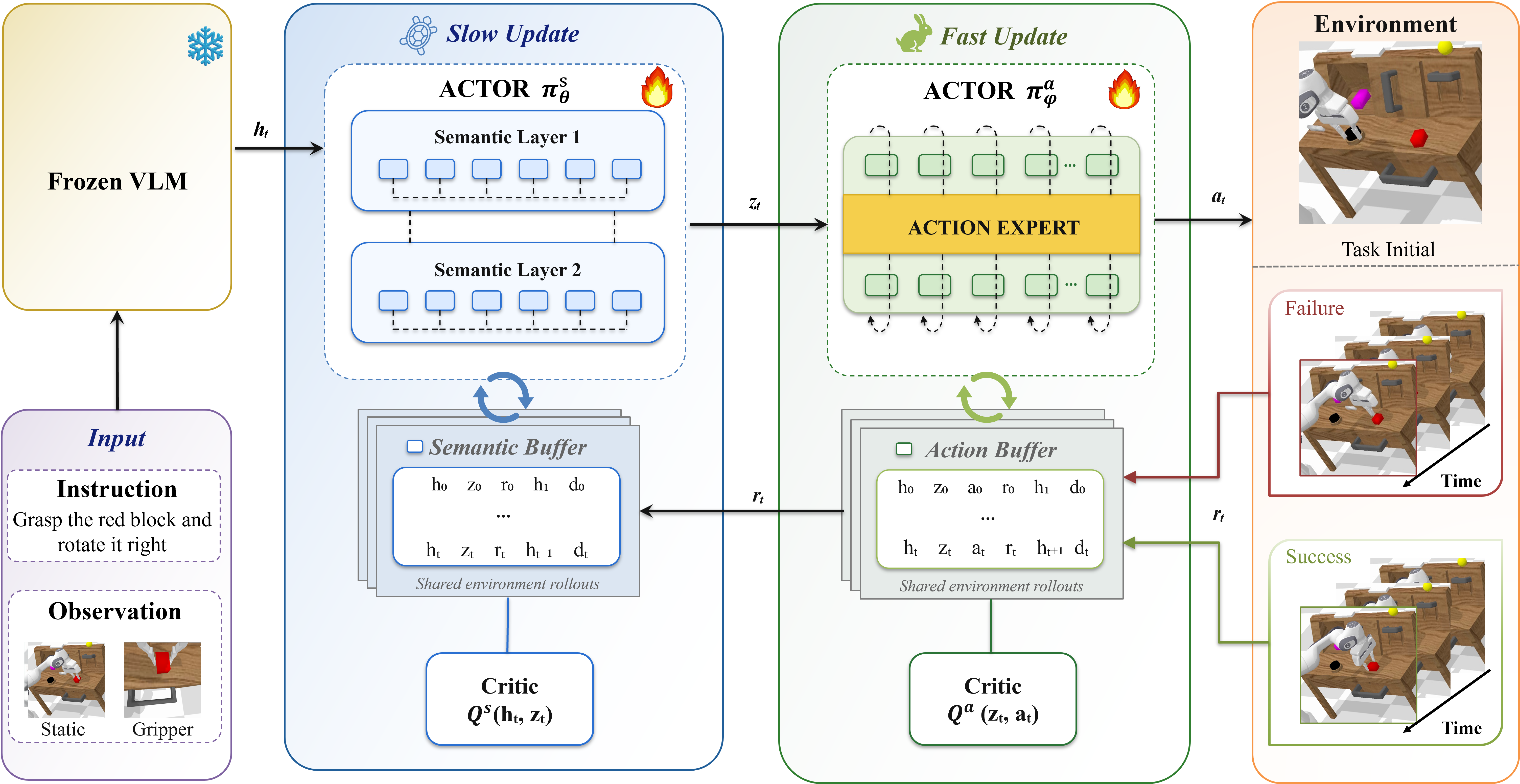}
\caption{\textbf{Framework of \methodname{}.} The frozen VLM encodes the instruction and observation into $h_t$, which is transformed by the semantic projection layer into the latent action $z_t$ and subsequently mapped by the action expert to the robot action chunk $\mathbf{a}_t$. Two separate module-level TD3 loops, each equipped with a dedicated replay buffer and critic, optimize the two modules using shared environment rollouts, with a lower update frequency for the semantic projection layer and a higher one for the action expert.}
\DIFaddbeginFL \vspace{-0.5cm}
\DIFaddendFL \label{fig:framework}
\end{figure*}
\subsection{Asynchronous VLA Execution and Optimization}

Acting While Understanding~\cite{yan2026acting} \DIFaddbegin \DIFadd{decouples inference computation: a low-frequency understanding module }\DIFaddend refreshes reusable semantic conditions, \DIFaddbegin \DIFadd{ while a high-frequency action module generates control actions from these conditions}\DIFaddend. RL-VLA$^3$~\cite{sun2026rlvla3} asynchronously coordinates environment interaction, model inference, and policy  \DIFaddbegin \DIFadd{updates to reduce waiting between training stages and }\DIFaddend improve system throughput.  \DIFaddbegin \DIFadd{Two-timescale stochastic approximation~\mbox{
\cite{borkar1997stochastic} }\hskip0pt
studies coupled updates with separated step-size schedules. Classical actor--critic algorithms~\mbox{
\cite{konda1999actor} }\hskip0pt
use this separation to let the critic track the value function while the policy adapts more slowly.
}

\DIFadd{TEMPO focuses on module-specific }\DIFaddend parameter-update frequencies  \DIFaddbegin \DIFadd{during VLA }\DIFaddend RL post-training\DIFaddbegin \DIFadd{, complementing work on inference efficiency, pipeline concurrency, and actor-critic timescales. It coordinates }\DIFaddend semantic-side and action-side  \DIFaddbegin \DIFadd{adaptation by updating the semantic projection layer less frequently than the action expert}\DIFaddend .

\section{\uppercase{Method}}
\label{sec:method}

\subsection{Problem Formulation}
\label{sec:problem_formulation}

We consider reinforcement learning post-training of a
pretrained VLA policy for robot
manipulation. The interaction between the robot and the
environment is modeled as a Markov decision process
$\mathcal{M}=(\mathcal{S},\mathcal{A},p,r,\gamma)$,
where $\mathcal{S}$ and $\mathcal{A}$ denote the state and
action spaces, $p$ denotes the
environment transition dynamics, $r$ is the task reward
function, and $\gamma$ is the discount factor. We formulate
the interaction at the action-chunk level, and $\mathcal{A}$
therefore denotes the action-chunk space. At each step $t$, the robot receives visual observation
$o_t$ and language instruction $l_t$. Given the state $s_t=(o_t,l_t)$, the policy generates an
action chunk $\mathbf{a}_t\in\mathcal{A}$, which is executed in
the environment. The environment provides a sparse reward only upon episode termination, denoted by $r_t$. The RL post-training objective is
\begin{equation}
\max_{\omega}
\mathbb{E}_{\tau\sim\pi_{\omega}}
\left[
\sum_{t=0}^{H-1}\gamma^t r_t
\right],
\label{eq:objective}
\end{equation}
where $\omega$ collectively denotes the trainable parameters of the VLA components selected for RL adaptation, $\tau$ denotes the closed-loop trajectory segment of the current active instruction, and $H$ is the number of policy decision steps before that instruction terminates.

\subsection{System Framework}
We instantiate the pretrained VLA policy with FLOWER~\cite{reuss2025flower}. As illustrated in Fig.~\ref{fig:framework}, the policy consists of a vision-language backbone $\mathrm{VL}$, a semantic projection layer $\pi_{\theta}^{s}$, and an action expert $\pi_{\phi}^{a}$. Given the visual observation $o_t$ and the language instruction $l_t$, the VLA policy is decomposed as
\begin{equation}
h_t=\mathrm{VL}(o_t,l_t),\qquad
z_t=\pi_{\theta}^{s}(h_t),\qquad
\mathbf{a}_t=\pi_{\phi}^{a}(z_t),
\label{eq:policy}
\end{equation}
where $h_t$ denotes the multimodal semantic representation, $z_t$ is the latent action provided to the action expert, and $\mathbf{a}_t$ is the robot action chunk.

We construct two separate module-level TD3~\cite{fujimoto2018addressing} learning loops around this architecture, each with its own replay buffer, twin critics, target networks, and optimizer. The framework decomposes RL post-training into two dedicated RL loops optimizing the semantic projection layer $\pi_{\theta}^{s}$
and the action expert $\pi_{\phi}^{a}$, while the pretrained vision-language backbone remains frozen. $\theta$ and $\phi$ denote the trainable parameters of $\pi_{\theta}^{s}$ and $\pi_{\phi}^{a}$,  \DIFaddbegin \DIFadd{collectively denoted by }\DIFaddend $\omega=(\theta,\phi)$.

For the \textit{semantic-level} RL loop, the state space $\mathcal{H}$ consists of the multimodal semantic representations $h_t$, while the action space $\mathcal{Z}$ consists of the latent actions $z_t=\pi_{\theta}^{s}(h_t)$. Accordingly, the semantic-level state-action critic $Q^{s}(h_t,z_t)$ evaluates and updates $\pi_{\theta}^{s}$.
For the \textit{action-level} RL loop, the state space is
$\mathcal{Z}$, with $z_t$ serving as the state, while the action space
$\mathcal{A}$ consists of the action chunks
$\mathbf{a}_t=\pi_{\phi}^{a}(z_t)$ executed in the environment. Similarly, the
action-level state-action critic $Q^{a}(z_t,\mathbf{a}_t)$ evaluates and updates
$\pi_{\phi}^{a}$.
Both loops share the  sparse reward $r_t=1$ if the current task  \DIFaddbegin \DIFadd{succeeds }\DIFaddend after executing $\mathbf{a}_t$, and $r_t=0$ otherwise.

The same interaction process generates the transition data required by both learning
loops. Each interaction transition is used to construct a semantic-level
transition for updating $Q^{s}$ and $\pi_{\theta}^{s}$, and an action-level
transition for updating $Q^{a}$ and $\pi_{\phi}^{a}$. 
The latent action $z_t$ produced by the semantic projection layer serves as the input to the
action expert, while the two modules are updated through their
respective critics and gradient pathways. This design allows the semantic projection layer and the action
expert to be optimized in their corresponding learning spaces
without direct gradient propagation between the two update
pathways. The semantic projection layer should remain comparatively stable to limit drift in the latent action space, whereas the action expert should adapt its control behavior in response to online interaction feedback. We therefore update the action expert at a higher frequency and the semantic projection layer at a lower frequency, as detailed in Section~\ref{sec:semantic-action frequency}.

\subsection{Semantic-Level TD3}

The semantic-level replay buffer stores
\begin{equation}
\mathcal{D}_S=
\left\{
(h_t,z_t,r_t,h_{t+1},d_t)
\right\},
\label{eq:semantic_buffer}
\end{equation}
where $d_t$ indicates whether the current active instruction
terminates after executing $\mathbf{a}_t$. Bellman bootstrapping~\cite{fujimoto2018addressing}  \DIFaddbegin \DIFadd{stops }\DIFaddend at each episode boundary, even when execution  continues with the next instruction in the long-horizon chain.

Let the twin semantic-level critics be denoted by~$Q_{1}^{s}$
and $Q_{2}^{s}$, their target networks by~$\bar{Q}_{i}^{s}$, and
the target semantic projection layer by $\bar{\pi}_{\theta}^{s}$.
The target latent action is generated as
\begin{equation}
\widetilde{z}_{t+1}
=
\bar{\pi}_{\theta}^{s}(h_{t+1}).
\label{eq:semantic_target}
\end{equation}

Using the target latent action, the semantic-level Bellman
target is defined as
\begin{equation}
y_t^S
=
r_t+\gamma(1-d_t)
\min_{i=1,2}
\bar{Q}_{i}^{s}
\left(
h_{t+1},\widetilde{z}_{t+1}
\right).
\label{eq:semantic_bellman_target}
\end{equation}

The twin semantic-level critics are optimized by minimizing
the squared Temporal-Difference (TD) error:
\begin{equation}
\mathcal{L}_{Q^{s}}
=
\sum_{i=1}^{2}
\mathbb{E}_{\mathcal{D}_S}
\left[
\left(
Q_{i}^{s}(h_t,z_t)-y_t^S
\right)^2
\right].
\label{eq:semantic_critic_loss}
\end{equation}

The semantic projection layer is optimized using the first
semantic-level critic:
\begin{equation}
\mathcal{L}_s
=
-
\mathbb{E}_{h_t\sim\mathcal{D}_S}
\left[
Q_{1}^{s}
\left(
h_t,\pi_{\theta}^{s}(h_t)
\right)
\right].
\label{eq:projection_loss}
\end{equation}
During this update, the replayed $h_t$ is treated as a fixed
input. Therefore, gradients are applied only to the semantic
projection parameters $\theta$ and do not propagate into the
frozen vision-language backbone or the action expert.

\subsection{Action-Level TD3}

The action-level replay buffer stores
\begin{equation}
\mathcal{D}_{A}
=
\left\{
\left(
h_t,
z_t,
\mathbf{a}_t,
r_t,
h_{t+1},
d_t
\right)
\right\},
\end{equation}
where $d_t$ follows the same active-instruction terminal definition as in the semantic-level replay buffer. The \DIFaddbegin \DIFadd{action-level critics use }\DIFaddend stored $z_t$ \DIFaddbegin \DIFadd{, while }\DIFaddend stored $h_t$ allows \DIFaddbegin \DIFadd{recomputation of }\DIFaddend the current latent action  \DIFaddbegin \DIFadd{for }\DIFaddend action-expert updates.

Let the twin action-level critics be denoted by $Q_{1}^{a}$ and
$Q_{2}^{a}$, their target networks by $\bar{Q}_{i}^{a}$, the target semantic
projection layer by $\bar{\pi}_{\theta}^{s}$, and the target action expert by
$\bar{\pi}_{\phi}^{a}$. The target latent action and
target action chunk are generated as
\begin{equation}
\tilde{z}_{t+1}
=
\bar{\pi}_{\theta}^{s}(h_{t+1}),
\qquad
\widetilde{\mathbf{a}}_{t+1}
=
\bar{\pi}_{\phi}^{a}(\tilde{z}_{t+1}).
\end{equation}

Using the target latent action and action chunk, the action-level
Bellman target is defined as
\begin{equation}
y^{A}_{t}
=
r_t
+
\gamma(1-d_t)
\min_{i=1,2}
\bar{Q}_{i}^{a}
\left(
\tilde{z}_{t+1},
\widetilde{\mathbf{a}}_{t+1}
\right).
\end{equation}

The twin action-level critics are optimized by minimizing the squared
temporal-difference error:
\begin{equation}
\mathcal{L}_{Q^{a}}
=
\sum_{i=1}^{2}
\mathbb{E}_{\mathcal{D}_{A}}
\left[
\left(
Q_{i}^{a}(z_t,\mathbf{a}_t)-y^{A}_{t}
\right)^2
\right].
\end{equation}

For the action-expert update, the current latent action is recomputed from the replayed $h_t$ as
$z_t^{\mathrm{sg}}
=
\operatorname{sg}\!\left(\pi_{\theta}^{s}(h_t)\right)$,
where $\operatorname{sg}(\cdot)$ denotes the stop-gradient operation.
The action expert is then optimized using the first action-level critic:
\begin{equation}
\mathcal{L}_{a}
=
-
\mathbb{E}_{h_t\sim\mathcal{D}_{A}}
\left[
Q_{1}^{a}
\left(
z_t^{\mathrm{sg}},
\pi_{\phi}^{a}(z_t^{\mathrm{sg}})
\right)
\right].
\end{equation}

During this update, $z_t^{\mathrm{sg}}$ is treated as a fixed input.
Therefore, gradients are applied only to the action-expert parameters
$\phi$ and do not propagate into the semantic projection layer or the
frozen vision-language backbone.

\subsection{Semantic-Action Frequency-Decoupled Optimization}
\label{sec:semantic-action frequency}
We further decouple the update frequencies of the semantic
projection layer and the action expert. Let $N_a$ and
$N_s$ denote the numbers of action-level and
semantic-level critic updates performed in each
reinforcement learning post-training round, respectively.
Both modules employ delayed actor updates. Specifically, the
action expert is updated every $d_a$ action-level critic
updates, whereas the semantic projection layer is updated
every $d_s$ semantic-level critic updates. Their effective
update frequencies per post-training round are therefore
\begin{equation}
f_a=\frac{N_a}{d_a},
\qquad
f_s=\frac{N_s}{d_s},
\label{eq:module_update_frequencies}
\end{equation}
where $f_a$ and $f_s$ denote the effective update
frequencies of the action expert and the semantic projection
layer, respectively. The action-to-semantic update-frequency
ratio is then defined as
\begin{equation}
\rho
=
\frac{f_a}{f_s}
=
\frac{N_a d_s}{N_s d_a}.
\label{eq:update_ratio}
\end{equation}
When $\rho=1$, the two modules are updated at the same frequency;
when $\rho>1$, the action expert is updated more frequently than
the semantic projection layer. Adjusting $\rho$ controls their
relative update frequencies. We find that settings with
$\rho>1$ generally outperform $\rho=1$, while different
frequency ratios yield different levels of improvement.
\section{\uppercase{Experiments}}
\label{sec:evaluation}

We  \DIFaddbegin \DIFadd{address four }\DIFaddend questions: (1) Does TEMPO improve long-horizon VLA performance over the pretrained state-of-the-art VLA policy and the RL post-training baseline? (2) Do the semantic projection layer and action expert provide complementary benefits during RL post-training? (3) How does the update-frequency ratio between the semantic projection layer and the action expert affect performance? (4) How does the post-training task-set size affect performance?

\subsection{Experimental Settings}

\textbf{Benchmark.} We evaluate on the CALVIN ABC$\rightarrow$D benchmark~\cite{mees2022calvin}, a long-horizon language-conditioned manipulation setting  \DIFaddbegin \DIFadd{with training }\DIFaddend on environments A, B, and C and  \DIFaddbegin \DIFadd{evaluation on }\DIFaddend unseen environment D. Each rollout contains a chain of five natural-language instructions.

\textbf{Metrics.} Following the standard CALVIN protocol, we report the success rates of completing the first $k$ tasks in a five-instruction chain, where $k=1,\ldots,5$. In Table~\ref{tab:sota}, columns 1-5 correspond to these consecutive success rates (SR). Avg. Len. is the average number of consecutively completed tasks from the beginning of each chain.

\textbf{Baselines.} \DIFadd{We compare our method with representative
language-conditioned multi-task policies on CALVIN, including
\textbf{GR-1}~\cite{wu2023gr1},
$\boldsymbol{\pi_0}$~\cite{black2025pi0},
$\boldsymbol{\pi_{0.5}}$~\cite{black2025pi05},
\textbf{UNIVLA}~\cite{wang2025univla}, and
\textbf{DeFI}~\cite{zhang2026defi}.
\textbf{FLOWER}~\cite{reuss2025flower} serves as the pretrained
initialization and reference policy.}
\textbf{FLOWER-RL}  \DIFaddbegin \DIFadd{is }\DIFaddend our single-loop  \DIFaddbegin \DIFadd{TD3 }\DIFaddend post-training baseline, \DIFaddbegin \DIFadd{ jointly updating }\DIFaddend the semantic projection layer and  action expert as a composite actor. In contrast, \textbf{\methodname{}} optimizes  \DIFaddbegin \DIFadd{them in separate }\DIFaddend TD3 loops  \DIFaddbegin \DIFadd{at independently controlled }\DIFaddend frequencies.

\begin{table}[t]
\vspace{0.2cm}
\centering
\caption{CALVIN ABC$\rightarrow$D experimental results.}
\label{tab:sota}
\scriptsize
\setlength{\tabcolsep}{0pt}
 \DIFaddbeginFL \renewcommand{\arraystretch}{1.08}
\newcommand{\srstd}[2]{\resizebox{0.123\columnwidth}{!}{\fontsize{7}{8}\selectfont#1$\pm$#2}}
\begin{tabular}{@{}>{\raggedright\arraybackslash}p{0.23\columnwidth}*{5}{>{\centering\arraybackslash}p{0.128\columnwidth}}>{\centering\arraybackslash}p{0.13\columnwidth}@{}}
\DIFaddendFL \toprule[1.0pt]
\textbf{Method} & \textbf{1} & \textbf{2} & \textbf{3} & \textbf{4} & \textbf{5} & \textbf{Avg. Len.} \\
\midrule
RoboFlamingo~\cite{li2023roboflamingo} & 82.4 & 61.9 & 46.6 & 33.1 & 23.5 & 2.47 \\
SuSIE~\cite{black2023susie} & 87.0 & 69.0 & 49.0 & 38.0 & 26.0 & 2.69 \\
GR-1~\cite{wu2023gr1} & 85.4 & 71.2 & 59.6 & 49.7 & 40.1 & 3.06 \\
OpenVLA~\cite{kim2025openvla} & 91.3 & 77.8 & 62.0 & 52.1 & 43.5 & 3.27 \\
CLOVER~\cite{bu2024clover} & 96.0 & 83.5 & 70.8 & 57.5 & 45.4 & 3.53 \\
RoboDual~\cite{bu2024robodual} & 94.4 & 82.7 & 72.1 & 62.4 & 54.4 & 3.66 \\
$\pi_0$~\cite{black2025pi0} & 93.8 & 85.0 & 76.7 & 68.1 & 59.9 & 3.84 \\
$\pi_{0.5}$~\cite{black2025pi05} & 92.7 & 84.6 & 77.3 & 69.3 & 61.2 & 3.85 \\
UP-VLA~\cite{zhang2025upvla} & 92.8 & 86.5 & 81.5 & 76.9 & 69.9 & 4.08 \\
RoboVLMs~\cite{li2026robovlms} & 98.0 & 93.6 & 85.4 & 77.8 & 70.4 & 4.25 \\
Seer-Large~\cite{tian2024seer} & 96.3 & 91.6 & 86.1 & 80.3 & 74.0 & 4.28 \\
VPP~\cite{hu2025vpp} & 96.5 & 90.9 & 86.6 & 82.0 & 76.9 & 4.33 \\
UNIVLA~\cite{wang2025univla} & 98.9 & 94.8 & 89.0 & 82.8 & 75.1 & 4.41 \\
DeFI~\cite{zhang2026defi} & 97.9 & 94.2 & 90.7 & 87.0 & 81.2 & 4.51 \\
\midrule
\rowcolor{resultgreen}
FLOWER~\cite{reuss2025flower} & 99.4 & 95.8 & 90.7 & 84.9 & 77.8 & 4.49 \\[3pt]
\rowcolor{resultgreen}
 FLOWER-RL & \textcolor{red}{\srstd{99.43}{0.35}} & \textcolor{red}{\textbf{\srstd{97.03}{0.58}}} & \textcolor{red}{\srstd{91.87}{0.74}} & \textcolor{red}{\srstd{85.37}{0.61}} & \textcolor{red}{\srstd{78.70}{0.70}} & \textcolor{red}{\srstd{4.52}{0.028}} \\[2pt]
\rowcolor{resultgreen}
 \textbf{\methodname{} (Ours)} & \textcolor{red}{\textbf{\srstd{99.83}{0.15}}} & \textcolor{red}{\srstd{96.77}{0.35}} & \textcolor{red}{\textbf{\srstd{92.77}{0.15}}} & \textcolor{red}{\textbf{\srstd{87.30}{0.20}}} & \textcolor{red}{\textbf{\srstd{81.70}{0.40}}} & \textcolor{red}{\textbf{\srstd{4.58}{0.005}}} \\[2pt]
\bottomrule[1.0pt]
\end{tabular}
\end{table}


\subsection{Simulation Results}

\textbf{Main Results.}  \DIFaddbegin \DIFadd{As shown in Table~\ref{tab:sota}, our }\DIFaddend method achieves the  \DIFaddbegin \DIFadd{highest SR5 among the listed methods }\DIFaddend under the CALVIN ABC$\rightarrow$D setting.  \DIFaddbegin \DIFadd{Across three seeds, TEMPO achieves }\DIFaddend an SR5 of  \DIFaddbegin \DIFadd{$81.70\pm0.40\%$, compared with $78.70\pm0.70\%$ for FLOWER-RL. The archived evaluation groups yield }\DIFaddend Avg. Len.  \DIFaddbegin \DIFadd{values of $4.58\pm0.005$ and $4.52\pm0.028$}\DIFaddend ,  respectively.  \DIFaddbegin \DIFadd{TEMPO improves }\DIFaddend SR5  \DIFaddbegin \DIFadd{by 0.50 percentage points over DeFI, the highest-SR5 baseline}\DIFaddend .

 \begin{table}[t]
\centering
\caption{Component ablation.}
\label{tab:component_ablation}
\scriptsize
\setlength{\tabcolsep}{3.0pt}
\begin{tabular*}{\columnwidth}{@{\extracolsep{\fill}}lcccccc@{}}
\toprule[1.0pt]
\textbf{Variant} & \textbf{1} & \textbf{2} & \textbf{3} & \textbf{4} & \textbf{5} & \textbf{Avg. Len.} \\
\midrule
Ours (full) & \textbf{100.0} & 97.1 & \textbf{92.9} & \textbf{87.1} & \textbf{81.7} & \textbf{4.59} \\
w/o projection update & 99.9 & 96.8 & 91.6 & 86.1 & 79.8 & 4.54 \\
w/o expert update & 99.9 & \textbf{97.2} & 92.2 & 85.6 & 79.6 & 4.55 \\
\bottomrule[1.0pt]
\end{tabular*}
\end{table}

\textbf{Component Ablation.} Table~\ref{tab:component_ablation} analyzes the contributions of the action expert and the semantic projection layer. The w/o projection update variant keeps the semantic projection layer fixed and updates only the action expert, whereas the w/o expert update variant keeps the action expert fixed and updates only the semantic projection layer. Both variants improve over the FLOWER reference policy, demonstrating that adapting either the semantic projection layer or the action expert can provide additional benefits under the sparse instruction-completion reward. However, neither variant reaches the performance of \methodname{}, which achieves the highest SR5 and Avg. Len. among all variants. These results indicate that the semantic projection layer and the action expert provide complementary adaptation capabilities. The similar performance of the two ablated variants suggests that neither module alone accounts for the full improvement.


\DIFaddbegin \textbf{\DIFadd{Frequency-Decoupling Analysis.}}
\DIFadd{Table~\ref{tab:update_ratio} evaluates action-to-semantic update-frequency ratios $\rho=f_a:f_s$, where $f_a$ and $f_s$ denote the effective update frequencies of the action expert and the semantic projection layer. Table~\ref{tab:optimization_budget} summarizes the configurations, each using 1M environment steps, 6,600 critic updates, and 3,300 actor updates.
}
\begin{figure*}[!t]
\centering
\includegraphics[width=\textwidth]{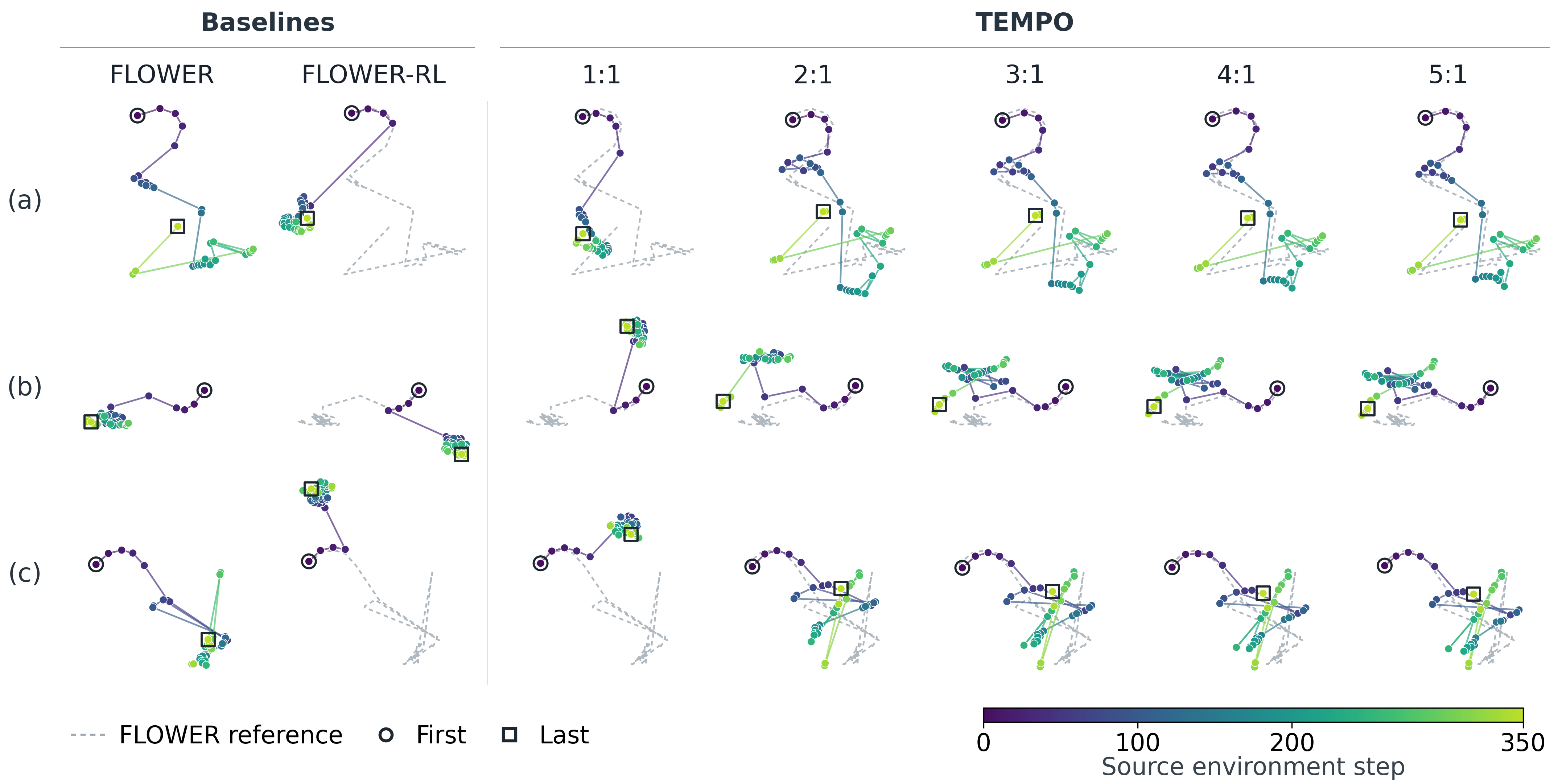}
\caption{\DIFaddFL{Latent-action comparisons on three FLOWER failure trajectories (rows) across FLOWER, FLOWER-RL, and TEMPO with update ratios of $1{:}1$--$5{:}1$ (columns). Models share 36 decision inputs per trajectory, spaced 10 environment steps apart. Within each row, normalized latent-action summaries are jointly reduced by PCA (1,024 to 50 dimensions), followed by t-SNE, with shared coordinates across columns. Colors indicate source environment steps, and solid lines connect consecutive inputs. Circles and squares mark the start and end; gray dashed lines show the FLOWER reference.}}
\label{fig:latent_action_drift}
\end{figure*}
\DIFaddend \begin{table}[t]
\vspace{0.2cm}
\centering
\caption{Action-to-semantic update-frequency ratios.}
\label{tab:update_ratio}
\scriptsize
\setlength{\tabcolsep}{3.0pt}
\begin{tabular*}{\columnwidth}{@{\extracolsep{\fill}}lcccccc@{}}
\toprule[1.0pt]
$f_a{:}f_s$ & \textbf{1} & \textbf{2} & \textbf{3} & \textbf{4} & \textbf{5} & \textbf{Avg. Len.} \\
\midrule
1:1 & 98.6 & 94.8 & 90.0 & 84.5 & 77.7 & 4.46 \\
5:1 & \textbf{100.0} & \textbf{97.1} & \textbf{92.9} & 87.1 & \textbf{81.7} & \textbf{4.59} \\
10:1 & 99.4 & 96.6 & 92.5 & \textbf{87.4} & 81.2 & 4.57 \\
\bottomrule[1.0pt]
\end{tabular*}
\end{table}

 \DIFaddbegin \begin{table}[t]
\color{red}\centering
\caption{\DIFaddFL{Update configurations.}}
\label{tab:optimization_budget}
\scriptsize
\setlength{\tabcolsep}{2pt}
\renewcommand{\arraystretch}{1.15}
\begin{tabular*}{\columnwidth}{@{\extracolsep{\fill}}lcccccc@{}}
\toprule[1.0pt]
\DIFaddFL{Ratio }& \shortstack{\DIFaddFL{Env.}\\\DIFaddFL{steps}} & \DIFaddFL{$N_a/N_s$ }& \DIFaddFL{$d_a/d_s$ }& \shortstack{\DIFaddFL{Critic updates}\\\DIFaddFL{(Action /}\\\DIFaddFL{Semantic)}} & \shortstack{\DIFaddFL{Actor updates}\\\DIFaddFL{(Action /}\\\DIFaddFL{Semantic)}} & \shortstack{\DIFaddFL{Optimization}\\\DIFaddFL{FLOPs}\\\DIFaddFL{(PFLOPs)}} \\
\midrule
\DIFaddFL{$1{:}1$ }& \DIFaddFL{1M }& \DIFaddFL{$33/33$ }& \DIFaddFL{$2/2$ }& \DIFaddFL{$3{,}300/3{,}300$ }& \DIFaddFL{$1{,}650/1{,}650$ }& \DIFaddFL{7.387 }\\
\DIFaddFL{$5{:}1$ }& \DIFaddFL{1M }& \DIFaddFL{$55/11$ }& \DIFaddFL{$2/2$ }& \DIFaddFL{$5{,}500/1{,}100$ }& \DIFaddFL{$2{,}750/550$ }& \DIFaddFL{12.276 }\\
\DIFaddFL{$10{:}1$ }& \DIFaddFL{1M }& \DIFaddFL{$60/6$ }& \DIFaddFL{$2/2$ }& \DIFaddFL{$6{,}000/600$ }& \DIFaddFL{$3{,}000/300$ }& \DIFaddFL{13.387 }\\
\bottomrule[1.0pt]
\vspace{-0.5cm}

\end{tabular*}
\end{table}

\DIFaddend \mbox{The $5{:}1$ and $10{:}1$ settings} improve SR5 over FLOWER by 3.9 and 3.4 points, respectively, whereas the equal-frequency 1:1 dual-loop setting is 0.1 point below FLOWER and reduces Avg. Len. from 4.49 to 4.46. These results indicate that  \DIFaddbegin \DIFadd{improvement depends on both RL updates and the modules' }\DIFaddend relative update frequencies. A relatively higher update frequency for the action expert enables it to incorporate interaction feedback more frequently, while less frequent updates to the semantic projection layer keep the latent action comparatively stable. The equal-frequency 1:1 setting removes this difference and weakens the overall gain.

\DIFaddbegin \textbf{\DIFadd{Latent-Action Drift Analysis.}} \DIFadd{Table~\ref{tab:latent_action_drift} compares latent-action drift and target-task success across different update-frequency ratios using checkpoints from the post-trained models at one million environment steps. We use three FLOWER failure trajectories on the target task }\texttt{\DIFadd{lift\_}\allowbreak{}\DIFadd{blue\_}\allowbreak{}\DIFadd{block\_}\allowbreak{}\DIFadd{slider}}\DIFadd{, retaining all 36 image--instruction decision inputs from each trajectory, for a total of 108 fixed inputs. For each input, we feed the same frozen-backbone features into each model's semantic projection layer.
}
\DIFaddbegin \begin{figure*}[!t]
\centering
\input{template/fig4_realworld_panels.tex}
\caption{\DIFaddFL{\realworldresultscaption}}
\label{fig:real_world_results}
\end{figure*}

\DIFadd{The semantic projection layer produces a latent action consisting of multiple 1,024-dimensional tokens. We average these token vectors and apply $L_2$ normalization to obtain a 1,024-dimensional latent-action summary $\hat{\mathbf{z}}$ for drift measurement and visualization. Drift is defined as the mean $L_2$ distance between each model's summary and the corresponding FLOWER summary under identical inputs:
}\begin{equation}
\DIFadd{D_{\mathrm{drift}} = \frac{1}{N}\sum_{i=1}^{N}
\left\|\hat{\mathbf{z}}_i^{\mathrm{trained}}-
\hat{\mathbf{z}}_i^{\mathrm{FLOWER}}\right\|_2.
\label{eq:latent_action_drift}
}\end{equation}
\DIFadd{Here, $N=108$ is the number of fixed inputs. We first average the distances within each trajectory, then report the mean and sample standard deviation across the three trajectory means. These deviations describe probe-trajectory variation, not variation across training seeds. Target-task success is evaluated over 50 trials from the same initial state, each limited to 360 environment steps.
}

\DIFadd{The $1{:}1$ configuration exhibits the largest drift, at 0.641, while FLOWER-RL also shows relatively large drift, at 0.611. In contrast, $4{:}1$ and $5{:}1$ reduce drift to 0.409 and 0.424, respectively, and achieve target-task success rates of 98.0\% and 96.0\%, both exceeding FLOWER-RL's 68.0\% and $1{:}1$'s 66.0\%. Figure~\ref{fig:latent_action_drift} visualizes the latent-action trajectories under these fixed inputs.
}

\DIFadd{These results support our proposed mechanism: less frequent updates to the semantic projection layer relative to the action expert help limit latent-action drift and provide more stable conditioning inputs, reducing the action expert's adaptation burden. Meanwhile, more frequent action-expert updates incorporate online interaction feedback to optimize control. This is consistent with the smaller drift and higher task success observed for $4{:}1$ and $5{:}1$.
}

\DIFaddend \begin{table}[t]
\color{red}\centering
\DIFaddbeginFL \caption{\DIFaddFL{Latent-action drift and target-task success.}}
\label{tab:latent_action_drift}
\footnotesize
\setlength{\tabcolsep}{5pt}
\begin{tabular*}{\columnwidth}{@{\extracolsep{\fill}}lcc@{}}
\toprule[1.0pt]
\textbf{\DIFaddFL{Model}} & \textbf{\boldmath\DIFaddFL{$D_{\mathrm{drift}}$}} & \textbf{\DIFaddFL{Success (\%)}} \\
\midrule
\DIFaddFL{FLOWER }& \DIFaddFL{$0.000 \pm 0.000$ }& \DIFaddFL{66.0 }\\
\DIFaddFL{FLOWER-RL }& \DIFaddFL{$0.611 \pm 0.041$ }& \DIFaddFL{68.0 }\\
\midrule
\DIFaddFL{$1{:}1$ }& \DIFaddFL{$0.641 \pm 0.034$ }& \DIFaddFL{66.0 }\\
\DIFaddFL{$2{:}1$ }& \DIFaddFL{$0.563 \pm 0.028$ }& \DIFaddFL{74.0 }\\
\DIFaddFL{$3{:}1$ }& \DIFaddFL{$0.470 \pm 0.028$ }& \DIFaddFL{72.0 }\\
\DIFaddFL{$4{:}1$ }& \DIFaddFL{$\mathbf{0.409 \pm 0.025}$ }& \textbf{\DIFaddFL{98.0}} \\
\DIFaddFL{$5{:}1$ }& \DIFaddFL{$0.424 \pm 0.027$ }& \DIFaddFL{96.0 }\\
\bottomrule[1.0pt]
\end{tabular*}
\end{table}

\begin{table}[t]
\centering
\DIFaddendFL \caption{RL POST-TRAINING TASK-SET SIZE COMPARISON.}
\label{tab:task_number}
\scriptsize
\setlength{\tabcolsep}{3.0pt}
\begin{tabular*}{\columnwidth}{@{\extracolsep{\fill}}ccccccc@{}}
\toprule[1.0pt]
\multicolumn{1}{l}{\textbf{RL task-set size}} & \textbf{1} & \textbf{2} & \textbf{3} & \textbf{4} & \textbf{5} & \textbf{Avg. Len.} \\
\midrule
1 & 99.4 & 96.6 & \textbf{92.5} & \textbf{87.4} & \textbf{81.2} & \textbf{4.57} \\
2 & 99.6 & \textbf{97.2} & 91.9 & 86.0 & 79.2 & 4.54 \\
4 & \textbf{99.9} & 96.9 & 92.0 & 86.8 & 80.1 & 4.56 \\
7 & 99.7 & 97.1 & 92.4 & 85.5 & 79.1 & 4.54 \\
34 & 99.6 & 96.8 & 92.2 & 86.3 & 79.2 & 4.54 \\
\bottomrule[1.0pt]
\end{tabular*}
\end{table}

\textbf{Post-Training Task-Set Size.} Table~\ref{tab:task_number} compares different RL task-set sizes under the fixed $10{:}1$ update-frequency ratio. Each rollout contains five subtasks, but only transitions associated with the selected task set are used for training. Thus,  task-set size  \DIFaddbegin \DIFadd{counts }\DIFaddend selected subtasks among CALVIN's 34 subtasks,  \DIFaddbegin \DIFadd{not }\DIFaddend instruction-chain length.

All evaluated settings outperform the FLOWER reference policy, with the single-task setting achieving the highest SR5 and Avg.~Len. However, performance does not improve monotonically as the task set expands: the full 34-task setting remains better than FLOWER but performs worse than the single-task setting. These results suggest that simply including more post-training tasks does not necessarily improve performance, and that the task composition and allocation of interaction data may require careful balancing.

\DIFaddbegin \textbf{\DIFadd{Architectural Generality.}} \DIFadd{TEMPO requires a frozen vision-language backbone and a trainable conditioning interface that can be optimized separately from the action expert. To assess applicability beyond FLOWER, we apply TEMPO to VLA-Adapter-Pro, using its native layer-wise key/value projections as the semantic projection module. Their outputs condition the corresponding blocks of a single action expert. The two modules use separate TD3 loops with an action-to-semantic update ratio of $5{:}1$.
}

\DIFadd{Table~\ref{tab:architectural_generality} reports means $\pm$ sample standard deviations across three seeds. TEMPO achieves a mean SR5 of 79.17\% and an Avg.~Len. of 4.487 on VLA-Adapter-Pro, compared with 77.23\% and 4.437 for the original model.
}

\begin{table}[t]
\color{red}\centering
\caption{\DIFaddFL{Architectural generality.}}
\label{tab:architectural_generality}
\scriptsize
\setlength{\tabcolsep}{2pt}
\newcommand{\archstat}[2]{\shortstack[c]{#1\\{\fontsize{6}{7}\selectfont$\pm$#2}}}
\begin{tabular*}{\columnwidth}{@{\extracolsep{\fill}}lrrrrrr@{}}
\toprule[1.0pt]
\textbf{\DIFaddFL{Model}} & \textbf{\DIFaddFL{SR1}} & \textbf{\DIFaddFL{SR2}} & \textbf{\DIFaddFL{SR3}} & \textbf{\DIFaddFL{SR4}} & \textbf{\DIFaddFL{SR5}} & \textbf{\DIFaddFL{Avg. Len.}} \\
\midrule
\DIFaddFL{VLA-Adapter-Pro }& \archstat{98.13}{0.61} & \archstat{94.60}{0.36} & \archstat{89.90}{0.26} & \archstat{83.80}{0.80} & \archstat{77.23}{1.31} & \archstat{4.437}{0.023} \\
\DIFaddFL{TEMPO }& \textbf{\archstat{98.37}{0.47}} & \textbf{\archstat{95.16}{0.51}} & \textbf{\archstat{90.92}{0.48}} & \textbf{\archstat{85.12}{0.35}} & \textbf{\archstat{79.17}{0.78}} & \textbf{\archstat{4.487}{0.009}} \\
\bottomrule[1.0pt]
\end{tabular*}
\end{table}

\DIFaddend \subsection{Real-World Experiments}
\label{sec:real_world}

\textbf{Experimental Setup.} We conduct multi-stage language-conditioned manipulation experiments on a physical platform comprising a 6-DoF Synria Alicia-D robot arm and a single-DoF gripper. Visual observations are provided by two Intel RealSense cameras: a D435i captures a third-person RGB view of the workspace, while a wrist-mounted D405 records close-range observations around the gripper. We evaluate  \DIFaddbegin \DIFadd{three }\DIFaddend multi-stage manipulation tasks. \DIFaddbegin \DIFadd{We }\DIFaddend collect 60 human demonstration episodes  \DIFaddbegin \DIFadd{per task, 180 in }\DIFaddend total. Each episode contains synchronized images, robot states, action commands, and language instructions. These demonstrations provide supervised data before online RL post-training.

\DIFaddend \textbf{Online RL Post-Training.} \DIFadd{The curves in Fig.~\ref{fig:real_world_results}(a)--(c)} report the evaluation reward during online RL post-training. For each random seed, we evaluate 20 trials at every checkpoint and sum binary task rewards, assigning 1 to success and 0 to failure. Across  \DIFaddbegin \DIFadd{three tasks, evaluation rewards initially fluctuate, }\DIFaddend then gradually improve and stabilize  \DIFaddbegin \DIFadd{during }\DIFaddend post-training. Compared with FLOWER-RL, \methodname{} reaches and maintains higher levels during later stages of post-training.

\textbf{Qualitative Analysis.}
\DIFadd{The top images in Fig.~\ref{fig:real_world_results}(a) and (b)
show initial task states and successful final states reached by \methodname{}}
when the drawer is not sufficiently open.
When the drawer is already fully open, FLOWER-RL and \methodname{}
achieve comparable performance. Their performance gap mainly emerges when
the limited drawer opening leaves insufficient space for direct target-object
manipulation. \DIFadd{In the recorded executions after} online RL post-training, \methodname{}  \DIFaddbegin \DIFadd{first opens
the drawer }\DIFaddend further and then completes the subsequent stages in the correct
order. In contrast,
FLOWER-RL directly attempts to manipulate the target object and fails.

\DIFaddbegin \DIFadd{The top images in Fig.~\ref{fig:real_world_qualitative_task3} show the initial task state
and a successful final state for the instruction ``take the car from the table and place it in the drawer.''
A yellow block initially rests on top of the car. In the recorded executions, FLOWER-RL manipulates
the block but does not complete the car transfer, whereas \methodname{}
first removes the block, then grasps the car and places it in the drawer.
}

\DIFadd{Overall, the higher late-stage evaluation rewards show that TEMPO improves task success more effectively than FLOWER-RL across the three real-world tasks. These results support the effectiveness of our semantic-action decoupled RL post-training framework for multi-stage manipulation.}

\DIFaddend 
\section{\uppercase{Conclusion}}
\label{sec:conclusion}


This paper presented \methodname{}, a semantic-action decoupled RL post-training framework that optimizes different VLA components through dedicated RL loops, adapting the semantic projection layer while keeping the pretrained vision-language backbone frozen, and optimizing the action expert to learn control- and execution-related behaviors from online feedback. A module-specific update-frequency strategy coordinates their adaptation, updating the semantic projection layer less frequently to mitigate drift in the latent action space and the action expert more frequently to incorporate control feedback efficiently. Evaluated on CALVIN and  \DIFaddbegin \DIFadd{three }\DIFaddend real-world manipulation tasks, \methodname{} achieves improved long-horizon performance over both state-of-the-art VLA models and the RL post-training baseline in simulation, as well as higher rewards in physical experiments.




\bibliographystyle{template/IEEEtran}
\bibliography{references}

\end{document}